\documentclass[letterpaper]{article} % DO NOT CHANGE THIS
\usepackage[submission]{aaai25}  % DO NOT CHANGE THIS
\usepackage{times}  % DO NOT CHANGE THIS
\usepackage{helvet}  % DO NOT CHANGE THIS
\usepackage{courier}  % DO NOT CHANGE THIS
\usepackage[hyphens]{url}  % DO NOT CHANGE THIS
\usepackage{graphicx} % DO NOT CHANGE THIS
\usepackage{subcaption}
\usepackage{natbib}  % DO NOT CHANGE THIS AND DO NOT ADD ANY OPTIONS TO IT
\usepackage{caption} % DO NOT CHANGE THIS AND DO NOT ADD ANY OPTIONS TO IT
\usepackage{newfloat}
\usepackage{listings}
\DeclareCaptionStyle{ruled}{labelfont=normalfont,labelsep=colon,strut=off} % DO NOT CHANGE THIS
\usepackage{booktabs}
\usepackage{multirow}
\usepackage{array}
\usepackage{multicol}

\usepackage{titling}
\usepackage{etoolbox}
\usepackage{amsmath} % 用于扩展数学功能
\usepackage{amssymb} % 提供更多数学符号
\usepackage{amsfonts} % 提供数学字体

\usepackage[linesnumbered,ruled,vlined]{algorithm2e} % 加载 algorithm2e 宏包

\usepackage{xcolor}
\newcommand{\mymodel}{DIPR\xspace}

\title{DIPR: Efficient Point Cloud Registration via Dynamic Iteration}

\author{Anonymous Authors}
\date{}
\author{
  \begin{tabular}{c} 
    Yang Ai$^{1}$ \and Qiang Bai$^{1}$ \and Jindong Li$^{1}$ \and Xi Yang$^{1,2,3,*}$
  \end{tabular} \\ \\
  \begin{tabular}{c}
    $^{1}$School of Artificial Intelligence, Jilin University, \\ 
    $^{2}$Engineering Research Center of Knowledge-Driven Human-Machine Intelligence, MoE, China,\\ 
    $^{3}$Key Laboratory of Ancient Chinese Script, Culture Relics and Artificial Intelligence, Jilin University \\
    aiyangjlu21@gmail.com, \{baiqiang23, jdli21\}@mails.jlu.edu.cn, yangxi21@jlu.edu.cn
  \end{tabular}
}

\begin{document}

\maketitle

\begin{abstract}
 % For point cloud registration, a significant challenge arises from non-overlapping points that consume extensive computational resources while negatively affecting registration accuracy. 
 % In this paper, we introduce a dynamic approach, widely utilized to improve network efficiency in computer vision tasks, to the point cloud registration task.
 % We employ this iterative registration process on point cloud data multiple times to identify regions where matching points cluster, ultimately enabling us to remove noisy points. 
 % Specifically, we begin with deep global sampling to perform coarse global registration. 
 % Then, we propose to use a refined nodes module to further narrow down the registration region and perform local registration. 
 % We also propose a spatial consistency-based classifier to evaluate the results of each registration stage so that our network terminates when sufficient confidence is reached, thus avoiding unnecessary computations.
 % Experiments show that our method improves registration accuracy while significantly reducing time and GPU memory consumption on multiple popular datasets, compared to the existing methods.

 Point cloud registration (PCR) is an essential task in 3D vision. 
 % Many existing methods have achieved satisfactory accuracy and precision. 
 Existing methods achieve increasingly higher accuracy.
 However, a large proportion of non-overlapping points in point cloud registration consume a lot of computational resources while negatively affecting registration accuracy. 
 To overcome this challenge, we introduce a novel Efficient Point Cloud Registration via Dynamic Iteration framework, \mymodel,
 that makes the neural network interactively focus on overlapping points based on sparser input points. 
 We design global and local registration stages to achieve efficient course-to-fine processing.
 Beyond basic matching modules, we propose the Refined Nodes to narrow down the scope of overlapping points by using adopted density-based clustering to significantly reduce the computation amount.
 And our SC Classifier serves as an early-exit mechanism to terminate the registration process in time according to matching accuracy. Extensive experiments on multiple datasets show that our proposed approach achieves superior registration accuracy while significantly reducing computational time and GPU memory consumption compared to state-of-the-art methods.
\end{abstract}

\section{Introduction}
\label{sec:intro}

Point cloud registration (PCR), which aims to align two or more point sets within the same coordinate system, plays a crucial role in many fields such as autonomous driving and robotics. With the advancement of deep learning, researchers are working on extracting learning-based features to replace un-robust traditional hand-crafted descriptors, e.g. GeoTrans~\cite{qin2022geometric}, RoITR~\cite{yu2023rotation}. However, due to the substantial data volume in practical scenarios, learning-based point cloud registrations tend to exhibit high time complexity and additional computational overhead. To address this problem, simple downsampling strategies, e.g., KPConv~\cite{thomas2019kpconv}, are employed in the existing methods, that select sparse points and their features from dense point clouds across hierarchical levels, and then pass them to an encoder for information interaction. 
However, this can result in numerous outliers (points in non-overlapping areas) during the registration process, which disrupt subsequent calculations for point feature pairing and unnecessarily consume significant computing resources, given that such downsamplings are typically global.

%------------------- Figure 1 ---------------------%
\begin{figure}[t]
  \includegraphics[width=1.0\linewidth]{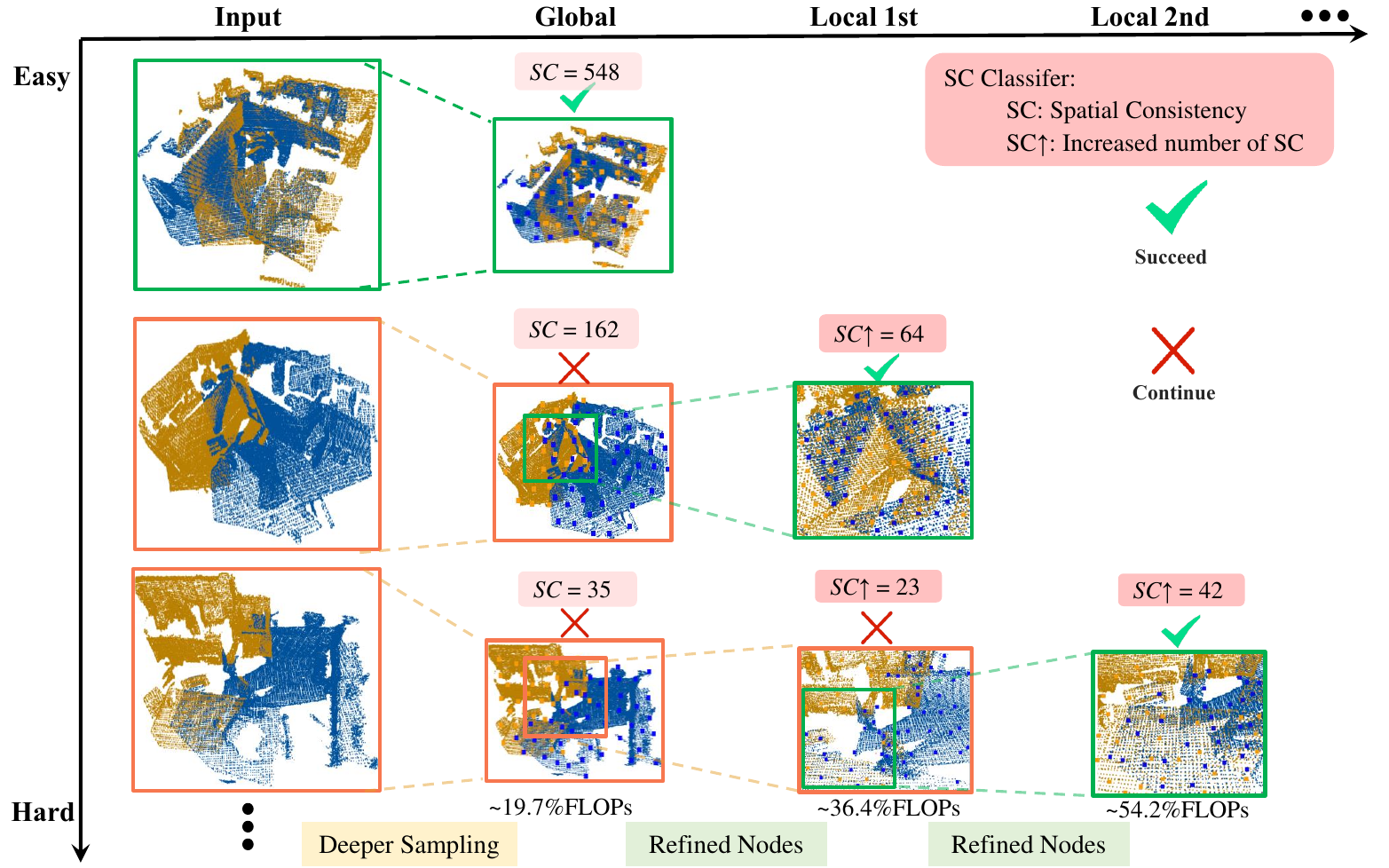}
  \caption{We propose a dynamic iteration approach to improve the accuracy and efficiency of PCR.
  % We propose a dynamic approach to iteratively conduct point cloud registration while improving accuracy and reducing computational consumption. 
  Our \mymodel samples sparser point clouds and dynamically search for overlapping areas in the local stage by Refined Nodes.
  After each stage, we design an SC Classifier to determine whether the registration result needs to be continued. Here, \(SC\) and \(SC{\uparrow}\) are the registration evaluation metrics.
  % $N_{geo}$, $N_{g}$ and $N_{l}^{i}$ \yang{do not use it before definition. A classifier is used to judge ...?} are the threshold used to judge whether the iteration is finished. 
  “FLOPs” refers to the proportion of the computation required by our method versus RoITR~\cite{yu2023rotation}.}% two stages!
\label{fig:teaser}
\vspace{-1.5em}
\end{figure}
%-----------------------------------------------%

% \ai{Dynamic networks are methods that can adjust network structures and parameters based on data. Starting from this observation, we introduce the spatial-wise dynamic network~\cite{wang2020glance} into PCR tasks to reduce computation resources during inference.}
Starting from this observation, we introduce the spatial-wise dynamic network~\cite{wang2020glance}, a dynamic network that can adjust network structures and parameters based on data, into PCR tasks to reduce computation resources during inference.
Although this approach was applied to 2D visual tasks, for 3D point clouds, the irregularity and larger scale of the data make managing spatial redundancy more challenging. Therefore, we propose a novel two-stage network with specifically designed Refined Nodes and spatial consistency-based SC Classifier to iteratively search regions within the inlier set, thus significantly reducing the computational cost of feature encoding.

As shown in Figure~\ref{fig:teaser}, our \mymodel primarily consists of two stages. First is the global registration stage, where we employ sparse global sampling for candidate pair selection and perform coarse registration with sparser inputs. The input scenes with discriminative features can easily be correctly aligned with high accuracy and plenty of pairings in this stage. Then, for complex scenes, the global registration may fail to produce sufficient correspondences, and our network proceeds to the second stage to select regions within the inlier set.
% In this local registration stage, we design a refined nodes module based on density-based clustering for the paired points from global registration to search regions where paired points are clustered together.
In this local registration stage, we design the Refined Nodes for the paired points from global registration. This module incorporates a density-based clustering algorithm and a Cluster-centric Neighborhood Augmentation (CNA) module. The clustering algorithm is used to identify regions where paired points are densely clustered, while the CNA searches within these regions to form data for the subsequent registration phase.
After each iteration, we also design a classifier based on spatial consistency to evaluate the registration performance based on dynamic thresholds and decide whether to proceed with the next iteration.

% Our proposed DIPR is different from other iteration-based point cloud studies (e.g., ~\cite{jiang2021planning,hosseini2018dynamic,yu2023peal}). Our main contributions could be summarized as follows:
Different from other iteration-based point cloud registrations ~\cite{jiang2021planning,yu2023peal} and baselines~\cite{wang2020glance, qin2022geometric, yu2023rotation}, our main contributions could be summarized as follows:

\begin{itemize}
    % \item novel network
    % \item refined nodes module
    % \item classifier
    % \item sota performance
    
    % \item We propose a novel registration network to dynamically discover the overlapping regions between point clouds, which effectively removes outliers and facilitates adaptive inference in the testing phase.
    % \item We develop a refined nodes module by using the DBSCAN algorithm to narrow down the point cloud registration region, efficiently reducing the computational load without introducing additional networks.
    % \item We achieve state-of-the-art performance on the 3DMatch and 3DLoMatch datasets while achieving time and computational efficiency advantages over other similar results.

    \item We propose a novel efficient point cloud registration via dynamic iteration framework, DIPR, to dynamically discover the overlapping regions between point clouds, which effectively removes outliers and facilitates adaptive inference in the testing phase. To the best of our knowledge, we are the first to introduce the dynamic approach to improve the efficiency of PCR.
    \item We design a Refined Nodes module with a clustering algorithm to narrow down the registration region, efficiently reducing the computational load without introducing additional networks. We also design a SC Classifier to serve as the early-exit mechanism of iterations. These structures enable our network to efficiently extract features using sparser input points (Deeper Sampling).
    \item We conduct comprehensive experiments on multiple public datasets (3DMatch, 3DLoMatch, KITTI), demonstrating our \mymodel achieves state-of-the-art performance while achieving time and computational efficiency advantages over existing methods.
\end{itemize}
%-------------------------------------------------------------------

\section{Related Work}
\label{sec:related}

\subsection{Correspondence-based Methods}
Traditional methods extract correspondences between two point clouds based on feature descriptors and predict the transformation matrix based on paired points~\cite{chen2022detarnet,xu2022finet}.
%and then calculate the transformation based on RANSAC~\cite{fischler1981random}. 
Recently, many variants~\cite{fragoso2013evsac, yang2020teaser} propose using outlier removal to improve the accuracy of correspondences. 
% DGR~\cite{choy2020deep} and 3DRegNet~\cite{pais20203dregnet} consider the inlier/outlier determination as a classification problem.
PointDSC~\cite{bai2021pointdsc} and SC$^{{2}}$-PCR~\cite{chen2022sc2} proposes the outlier rejection module by introducing deep spatial consistency and global consistency to measure the similarity between correspondences. 
% Recently, MAC~\cite{zhang20233d} developed a compatibility graph to render the affinity relationship between correspondences and construct the maximal cliques to represent consensus sets. 
Although these methods~\cite{shen2022reliable,huang2024efficient} have made significant progress, they are all greatly affected by initial correspondence. In comparison, our work insights by the idea of dynamically selecting matching points to improve the accuracy of correspondences.
% \vspace{-0.3cm}
\subsection{Deep-learned Feature Descriptors} 
The feature descriptor in the PCR is used to extract features to construct correspondences. Compared with traditional hand-crafted descriptors such as~\cite{rusu2009fast, chu2011multi}, 
%descriptors based on deep learning have developed rapidly in recent years. As a pioneer, 
3DMatch~\cite{zeng20173dmatch} learns 3D geometric features to convert local patches into voxel representations. 
%Later, PointNetLK~\cite{aoki2019pointnetlk} and PCRNet~\cite{sarode2019pcrnet} integrate PointNet~\cite{qi2017pointnet} into the point registration task and other deep learning methods~\cite{deng2018ppfnet, yew2020rpm} improve feature extraction module to enhances the capability of feature descriptors. 
% However, these methods are greatly affected by the overlap rate. 
In recent years, more and more works~\cite{zhang2022end, yu2022riga} focus on the low overlap problems. Predator~\cite{huangpredator} extracts low-overlap point clouds dataset to form 3DLoMatch and uses the cross-attention module to interact with the point clouds to build feature descriptors. 
% CoFiNet~\cite{yu2021cofinet} leverages a two-stage method, using down-sampling and similarity matrices for rough matching and up-sampling the paired point neighborhood to obtain fine matching. 
GeoTrans~\cite{qin2022geometric} and other papers~\cite{yu2023rotation} learn more geometric features to improve the robustness of PCR and achieves high accuracy. 
% Additionally, RoITR~\cite{yu2023rotation} addresses rotation-invariant problems by introducing novel local attention mechanisms and global rotation-invariant transformers. 
These methods compute features for each point and its neighborhood, resulting in an unavoidably large amount of computation. 
% BUFFER~\cite{ao2023buffer} attempts to improve the computation efficiency, but its performance is mainly reflected in its generalization ability. 
Recently, many of the methods have used additional information to improve the PCR process, such as~\cite{yu2023peal, chen2024dynamic, chen2024pointreggpt}. 

However, these methods are challenging to implement in real-world scenarios due to their complexity and high resource demands.
In comparison, our method uses sparser sampling in the down-sampling stage to reduce the consumption of computational resources.
% \vspace{-0.3cm}
\subsection{Dynamic Neural Networks} 
The dynamic neural network is a structure that dynamically changes based on computational resources and input data. 
% Compared to static networks, dynamic networks can adapt parameters or structures to different inputs, leading to notable advantages in terms of accuracy, computational efficiency, adaptiveness, etc~\cite{han2021dynamic}. 
% In recent years, many advanced dynamic network works have been proposed and mainly divided into three different aspects: 1) sample-wise dynamic networks~\cite{yang2019condconv} are designed to adjust network architectures and parameters to reduce computation; 2) temporal-wise dynamic networks~\cite{meng2020adafuse} treat video data as a sequence of images to save computation along temporal dimension; 
% 3) spatial-wise dynamic networks~\cite{zhou2016learning} exploit spatial redundancy to improve efficiency. 
Our approach is mainly related to spatial-wise~\cite{zhou2016learning} dynamic networks which exploit spatial redundancy to improve efficiency. Mobilenet~\cite{howard2017mobilenets} observes that a low resolution might be sufficient for most ”easy” samples and exploits feature solutions to remove redundancy. 
Besides, early-exiting mechanism~\cite{li2019improved,yu2023boosted} is widely studied and used to decide whether the network continues. 
GFNet~\cite{wang2020glance} imitates the visual system of the human eye and processes an observation process from a global glance to a local focus to search more critical regions. Based on GFNet, we propose a coarse-to-fine feature learning approach combined with registration evaluation to adjust the registration range. 

%----------------- Figure: Framework ----------------%
\begin{figure*}[t]
  \centering
  \includegraphics[width=0.95\textwidth]{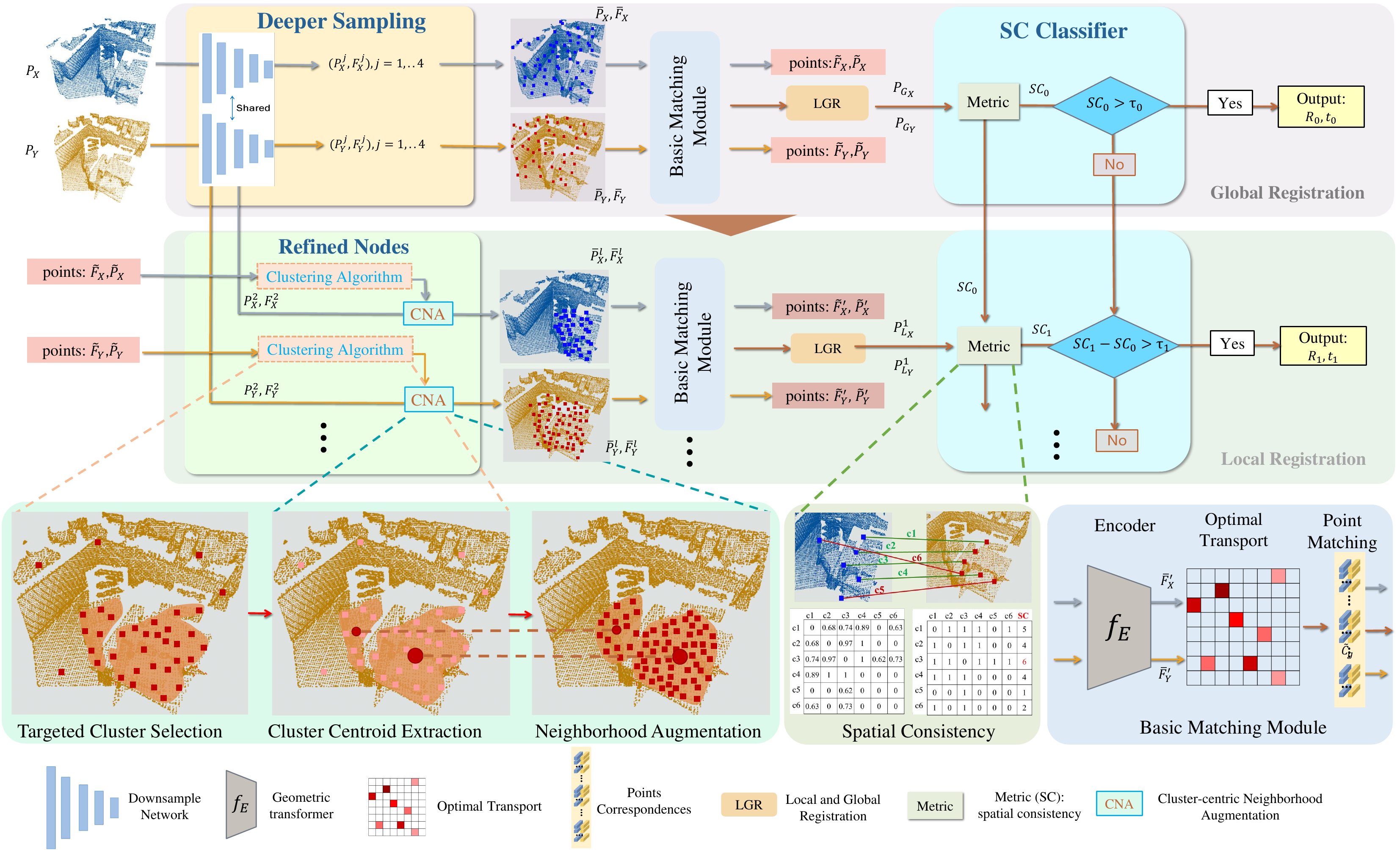}
  % \vspace{-1em}
   \caption{  
   Our \mymodel is mainly divided into two parts: global registration and local registration. We first design the Deeper Sampling to downsampling and extract the initial feature. Then, we use the Basic Matching Module to obtain the pairing points \(\tilde{P}_{X}\) and \(\tilde{P}_{Y}\) and the global registration result. Next, the Refined Nodes in the local registration stage clusters the paired points and obtains the new coarse node for local registration. After each registration stage finishes, we design the SC Classifier to determine whether the current result requires further iteration.}
  \label{fig:pipline}
  \vspace{-1em}
\end{figure*}
%------------------------------------------------%

%%%%%%%%%%%%%%%%%%%%%%%%%%%%%%%%%%%%%%%%%%%%%%%%%%%%%%%
\section{Method}
\label{sec:method}

Given a pair of 3D point clouds \(P_{X}=\{p_{x_i}\in \mathbb{R}^3 \,|\,i = 1,\ldots, N \}\) and \(P_{Y}=\{p_{y_i}\in \mathbb{R}^3 \,|\,i = 1,\ldots, M \}\), the purpose of our task is to estimate a rigid transformation $SE(3)$ which aligns the point clouds X and Y with a rotation matrix $R \in SO(3)$ and a translation $T \in \mathbb{R}^3$.
Figure~\ref{fig:pipline} shows the overview of our method. 

Our method is divided into two parts, namely the global registration stage and the local registration stage. In the global stage, we first employ the Deeper Sampling to perform multi-scale deep sampling on the input point clouds and extract initial features of each point.
Basic Matching module is employed to conduct coarse-grained registration of the data. 
For the local registration stage, we employ the Refined Nodes module to get the overlapping area based on the previous registration result and start the next registration. 
After every registration stage, the SC Classifier is designed to judge whether the results meet the accuracy requirements.
%%%%%%%%%%%%%%%%%%%%%%%%%%%%%%%%%%%%%%%%%%%%%%%%%%%%%%%%%%%%
\subsection{Global Registration}
\label{sec:global registration}
% We first pass the sampled point cloud through the encoder to extract geometric features, and pair them based on the similarity of the features. The points matching module is then used to form more pairing relationships around the paired nodes, and calculate the global registration result. 
% \ai{We design \textbf{Deeper Sampling} for global downsampling and feature extraction of the entire point cloud. Subsequently, the Basic Matching module extracts geometric features from the point clouds and establishes correspondences based on similar features. Finally, we design LGR to compute the global registration results.}

In the global registration stage, we aim to align the entire point cloud by using a hierarchical approach. This process involves Deeper Sampling for feature extraction, Basic Matching to establish initial correspondences, and Local-to-Global Registration (LGR) to refine these correspondences, ensuring accurate and computationally efficient alignment.

\paragraph{\textbf{Deeper Sampling.}}
\label{sec:sparse_sense}
We perform deeper uniform downsampling in the original input data. Inspired by GFNet~\cite{wang2020glance}, we observe that in point cloud registration tasks, full data may not always be necessary for computations and that only partially accurate pairing relationships can yield satisfactory results. In the global registration stage, we use a minimal number of data points to create a global representation of the point cloud for the global registration, reserving more computational resources for the subsequent local registration phase.
We utilize the downsampling network to sample the original point clouds \(P_{X}\) and \(P_{Y}\) to obtain multiple levels of points \(P_{X}^{j}\in \mathbb{R}^ {N_{j}*3}\), \(P_{Y}^{j}\in \mathbb{R}^{M_{j}*3}\) and their corresponding features \(F_{X}^{j}\in\mathbb{R}^{N_{j}*D}\), \(F_{Y}^{j}\in \mathbb{R}^{M_{j}*D}\) ($j$ represents the number of sampling layers). Our approach allows using different networks for downsampling. In this work, we employ two SOTA backbones, KPConv~\cite{thomas2019kpconv} and the Global Aggregation module from RoITR~\cite{yu2023rotation}, to demonstrate the effectiveness of our method. Compared with KPConv, Global Aggregation (GA) takes the point cloud coordinates and estimated normals as input, and hierarchically downsamples the original point cloud to coarser nodes through a set of encoder-decoder modules. Due to the addition of normal information, it has better geometric feature representation capabilities, but it also increases the computational complexity. Compared to previous PCR approaches~\cite{yu2021cofinet, qin2022geometric, yu2023rotation}, we employ a deeper level of sparse downsampling (4 times) during the downsampling process to achieve a sparser global representation, significantly reducing our computational load. We differ from the existing methods in that we use an iterative registration approach. During the subsequent local registration process, we re-obtain coarse nodes around the inliers to avoid reducing the accuracy of the registration results due to the sparse sampling.

%%%%%%%%%%%%%%%%%%%%%%%%%%%%%%%%%%%%%%%%%%%%%%%%%%%%%%%%%%%%%%%%%%%%%%%%%%
\paragraph{\textbf{Basic Matching Module.}}
\label{sec:basic matching}
We pass the sparse sampled point cloud \(\Bar{P}_{X}\), \(\Bar{P}_{Y}\)  along with its corresponding features \(\Bar{F}_{X}\), \(\Bar{F}_{Y}\) as coarse nodes into the encoder to learn background geometric information~\cite{qin2022geometric}. The encoder consists of a geometric embedding submodule and multiple transformers. Each transformer includes a self-attention structure, a cross-attention structure and a feedforward network. This process enhances the feature similarity between the two point clouds while improving the feature representation capability. After obtaining the coarse node features \(\Bar{F}_{X}^{'}\),\(\Bar{F}_{Y}^{'}\), we perform the optimal transport to normalize the node features and measure the pairwise similarity to select the top-\({k}\) as the coarse correspondence set \(C = \{(x_{i}^{'},y_{j}^{'})\mid x_{i}^{'}\in \Bar{P}_{X},y_{j}^{'}\in \Bar{P}_{Y}\} \).

We employ a coarse-to-fine point matching module to upsample the correspondence coarse nodes to a denser point set, forming patches \(\tilde{P}_{X} \subseteq P_{X}^{2}\),\(\tilde{P}_{Y} \subseteq P_{Y}^{2}\) around the nodes while preserving the pairing relationships between them. Within these corresponding patches, we perform point matching based on feature \(\tilde{F}_{X} \subseteq F_{X}^{2}\),\(\tilde{F}_{Y} \subseteq F_{Y}^{2}\), resulting in additional pairwise relationships at the point level, denoted as \(\tilde{C} = \{(\tilde{x}_{i},\tilde{y}_{j})\mid \tilde{x}_{i}\in \tilde{P}_{X},\tilde{y}_{j}\in \tilde{P}_{Y}\} \), and the similarity \(\tilde{S} = \{\tilde{s}_{i}\mid \tilde{s}_{i}\in \mathbb{R}^{|\tilde{C}|*1}\}\) for each pair, which will be further used in the subsequent local registration process.

\paragraph{\textbf{Local-to-Global Registration (LGR).}}
\label{{sec:LGR}}
We use  LGR to register the point cloud. LGR first calculates transformations for each patch using its local point correspondences:
\begin{equation} 
R_{i}, t_{i} = \min_{R, t}\textstyle \sum_{(\tilde{x}_{i}, \tilde{y}_{i}) \in \tilde{C}_{i}} \tilde{w}_{i}||R\cdot \tilde{x}_{i} + t -\tilde{y}_{i}|| _{2}^{2} 
\end{equation}
We can obtain \(T_{i} = \{R_{i}, t_{i}\}\) by using weighted SVD to solve the above equation in closed form. The match similarity score for each correspondence in \(\tilde{S}\) is used as \(\tilde{w}_{i}\). In all patch transformations, we select the best one as the global transformation and iteratively filter paired points over the global paired points.
\begin{equation} R, t = \max_{R_{i}, t_{i}}\textstyle \sum_{(\tilde{x}_{i}, \tilde{y}_{i}) \in \tilde{C}} \left[||R_{i}\cdot \tilde{x}_{i} + t_{i} -\tilde{y}_{i}||_{2}^{2} < \tau_{t} \right] \end{equation}
where \(\left[ \cdot \right]\) is the Iverson bracket, \(\tau_{t}\) is the acceptance threshold. After N iterations, we obtain the final pairing relationship \(\Hat{C} = \{(\Hat{x}_{i},\Hat{y}_{j})\mid \Hat{x}_{i}\in \tilde{P}_{X},\Hat{y}_{j}\in \tilde{P}_{Y}\} \).

%%%%%%%%%%%%%%%%%%%%%%%%%%%%%%%%%%%%%%%%%%%%%%%%%%%%%%%%%%%%%%%%
\subsection{Local Registration}
\label{sec:local registration}
After global registration, we use Refined Nodes to obtain new coarse nodes based on the points pairing relationship. In addition, the SC Classifier is used to evaluate the registration results to determine whether to continue after each registration.

\paragraph{\textbf{Refined Nodes.}} 
\label{sec:Refined Nodes Module}
This module primarily consists of three components: Targeted Cluster Selection, Cluster Centroid Extraction, and Neighborhood Augmentation. During the local registration stage, we aim to narrow down the data scope to reduce computational complexity while avoiding interference from outliers. For pairing relationships \(\tilde{C}\), where the paired points \(\tilde{x}_{i}\) and \(\tilde{y}_{i}\) represent successfully matched points in these two point clouds \(\tilde{P}_{X}\) and \(\tilde{P}_{Y}\), and the pairing similarity \(\tilde{S}\) indicates the degree of similarity between each pair of relationships, which remove a significant portion of point cloud data that could not be paired compared to the original globally sampled point set. 
The Target Cluster submodule employs an adaptive clustering method, based on the density-based clustering method (DBSCAN) ~\cite{ester1996density}, to combine paired similarity \(\tilde{S}\) and the relative distances within the point sets \(\tilde{P}_{X}\) and \(\tilde{P}_{Y}\) for clustering.
% We use the density-based clustering method (DBSCAN) ~\cite{ester1996density} combined with the pairing similarity \(\tilde{S}\) to cluster the \(\tilde{P}_{X}\) and \(\tilde{P}_{Y}\) to find the region \(P_{X}^{d}\) and \(P_{Y}^{d}\) where inliers are clustered. 
The specific implementation is as shown in Algorithm~\ref{alg:refined}.

%--------------------- pseudo code ---------------------%
\begin{algorithm}[t]
% \small
\scriptsize
% \tiny
\SetKwInOut{Input}{Input}
\SetKwInOut{Output}{Result}
\SetKwInOut{Parameters}{Parameters}
\SetKw{Require}{Require:}

\Input{$P$ - Points cloud data, label - Points labels, $W$ - Points weights}
\Parameters{$\epsilon$ - Radius, $minPts$ - Density threshold, dist - Distance function} 
\Output{cluster - Cluster points, $M$ - Clusters Num}
\caption{Our Clustering Algorithm Pseudocode}
\label{alg:refined}

CountClusters $C \leftarrow 0$\;
\While{True}{
    Distance $D \leftarrow$ dist($P$, $W$)\;
    \ForEach{point $p$ in $P$}{
        \If{label($p$) is undefined}{
            Neighbors $N \leftarrow$ RangeQuery($D$, $p$, $\epsilon$)\;
            \If{$|N| < \text{minPts}$}{
                label($p$) $\leftarrow$ Noise\;
            }
            \Else{
                $c \leftarrow$ NextClusterLabel\;
                ExpandCluster($D$, $N$, $c$, $\epsilon$, minPts)\;
            }
        }
    }
    ClustersNum $M \leftarrow$ UniqueLabels(label)\;
    \If{$M > C$}{
        $C \leftarrow M$, $\epsilon \leftarrow \epsilon \cdot 1.1$, minPts $\leftarrow$ minPts $+ 1$\;
    }
    \Else{
        \textbf{break}\;
    }
}
cluster $\leftarrow$ ExtractCluster($P$, label)\;
\end{algorithm}
%---------------------------------------------------%
By combining \(\tilde{S}\), we can exclude those points with low pairing similarity. 
Subsequently, in the Cluster Centroid Extraction submodule, we calculate the cluster centers of the clustered point sets \(P_{X}^{d}\) and \(P_{Y}^{d}\) to obtain the cluster centers. We compute the distances from each point in the original point set \(P_{X}^{2}\) and \(P_{Y}^{2}\) to cluster centers and sort them. By ranking the distances, we select the top \(N_{r}\) closest points to achieve Neighborhood Augmentation for the cluster centers. To ensure the efficiency of the local registration stage and the accuracy of the registration results, we maintain the number of refined nodes \(N_{X}^{r}\), \(N_{Y}^{r}\) equal to the number of anchor points \(\Bar{P}_{X}\), \(\Bar{P}_{Y}\) used in the global registration. \(\Bar{P}_{X}^{l}\), \(\Bar{P}_{Y}^{l}\) are then used as the next set of coarse nodes for local registration.
% Subsequently, we compute the cluster centers for the clustered point sets \(P_{X}^{d}\) and \(P_{Y}^{d}\) and map them back to the original point set \(P_{X}^{2}\) and \(P_{Y}^{2}\) using K-nearest neighbors (KNN) as our next set of coarse nodes \(P_{X}^{c}\) and \(P_{Y}^{c}\) for local registration. 
The process of getting \(P_{X}^{c}\) is shown:
\begin{equation} 
K(P_{X}^{d},k) = \{p_{x_{i}}^{l}| p_{x_{i}}^{l} \in P_{X}^{2} , dist(p_{x_{i}}^{d}, p_{x_{i}}^{l}) \le d_{k}\}
\end{equation}

Compared to the coarse nodes from the earlier global registration, nodes refined in this manner are more tightly clustered. This refined process also eliminates a significant number of outliers, reducing their impact on the results and computational resource utilization.

\paragraph{\textbf{Local Matching.}} 
The same Basic Matching module is used for the local registration stage. However, this encoder shares the same network architecture as the encoder used in the global stage, but with different parameters. While the global encoder learns sparse global information, the local encoder focuses on learning denser local regions. Using the same encoder for both would lead to a significant decline in network performance due to their differing tasks. The point matching submodule and LGR module remain consistent with those used in the global registration process.
% Similar to the coarse node encoder, the local encoder is used to extract deep representations of nodes. They share the same network structure but with different parameters. In the global registration, the encoder learns sparse global information, while the local encoder processes denser local regions for learning. If the same network were used, the differences between them would result in a noticeable reduction in the performance of the network.For point matching submodule and registration calculations remain consistent with the global registration process.

%%%%%%%%%%%%%%%%%%%%%%%%%%%%%%%%%%%%%%%%%%%%%%%%%%%%%%%%%%%%%%%%%%%%%%%%%%
% \subsection{SC Classifier}
% \label{sec:classifier and loss}
% After each registration stage is completed, we use the classifier to determine whether the registration result should continue to the next iteration. The classifier serves as the early-exit mechanism to reduce unnecessary computing consumption and dynamically allocate computing resources.

\paragraph{\textbf{SC Classifier.}}
\label{sec:classifier and loss}
After each registration stage is completed, we use the classifier to determine whether the registration result should continue to the next iteration. The classifier serves as the early-exit mechanism to reduce unnecessary computation and dynamically allocate computing resources.

Compared to visual tasks such as object recognition, point cloud registration tasks are challenging to employ deterministic metrics during the inference phase to ascertain whether the current result has achieved the optimal state. We empirically employ spatial consistency \(SC\) measurement~~\cite{bai2021pointdsc} to assess the level of accuracy achieved in the current registration process, allowing us to evaluate the results of each registration attempt and determine whether they have achieved a high level of accuracy. For the pairing relationship \(\Hat{C}\), the SC measure is defined:
\begin{equation}
SC_{i,j} = \Phi(d_{i,j}) = [1 - \frac{d_{i,j}^{2}}{\sigma^2_{d}}]_{+}, d_{i,j} = |d(\Hat{x}_{i},\Hat{x}_{j})-d(\Hat{y}_{i},\Hat{y}_{j})
\end{equation}
where \(\Phi(\cdot)\) is a monotonically decreasing kernel function. $ \left [ \cdot \right ] _ {+}$ is the max$(\cdot, 0)$ operation. \(d(\cdot,\cdot)\) is the Euclidean distance between two points and $\sigma_{d}$ is a distance parameter. \((\Hat{x}_{i},\Hat{x}_{j})\) and \((\Hat{y}_{i},\Hat{y}_{j})\) are the \(i, j\) paired points of correspondence \(\Hat{C}\).  
For the global registration stage, we set the global \(SC\) threshold \(N_{g}\) to determine whether the registration can end. For the local registration stage, we set the local \(SC\) comparison threshold \(N_{l}^{i}\) to compare whether the result of the current local registration is better than the previous iteration. If the result of SC is less than \(N_{ l}^{i}\), it is considered that the results have deteriorated during the iteration process and need to be terminated, otherwise, the next local registration will continue. Using this comparison threshold as the end requirement helps to dynamically determine whether the iteration is over while avoiding the problem of insufficient versatility in different scenarios using a fixed threshold. 

The SC Classifier allows the algorithm to promptly end the iteration process without excessive resource consumption. This helps improve efficiency and achieve better registration results within limited computational resources. During the network training process, we inactivate early-exit mechanism by not using the classifier. We assume that before reaching the set maximum number of iterations, each registration result requires the next iteration to ensure that the performance of the network reaches its optimum.

%%%%%%%%%%%%%%%%%%%%%%%%%%%%%%%%%%%%%%%%%%%%%%%%%
\paragraph{\textbf{Loss Function.}}
Our loss consists of two components: \(L = L_{c} + \lambda L_{p}\), with a coarse matching loss \(L_{c}\) and a point matching loss \(L_{p}\) are derived from the registration baseline~\cite{qin2022geometric}. The loss weight \(\lambda\) is used to balance the importance of different loss functions. For a detailed definition, please refer to our supplementary material.

%------------------------------------------------------------------------
%%%%%%%%%%%%%%%%%%%%%%%%%%%%%%%%%%%%%%%%%%%%%%%%%%%%%%%%%%%%%%%%%%%%%%%%%%
\section{Experiments}
\label{sec:experiments}

\subsection{Indoor Benchmarks: 3DMatch and 3DLoMatch}
\label{sec:3d}
\paragraph{\textbf{Dataset and Metrics.}} 
We utilize two popular datasets to evaluate the real-world performance of our method as well as other methods. 
The 3DMatch~\cite{zeng20173dmatch} dataset is preprocessed and divided into two categories based on the level of overlap: 3DMatch (\textgreater\ $30\%$ overlap) and 3DLoMatch ($10\%$ $\sim$ $30\%$ overlap). 
We use Registration Recall (RR) (most important), Relative Translation Errors (RTE) and Relative Rotation Errors (RRE) to measure the accuracy of registrations. More implementation details are described in our supplementary material.

\begin{figure}[t]%{0.5\textwidth}
\centering
    \begin{minipage}[l]{0.24\textwidth}
        \centering
        \includegraphics[width=\textwidth]{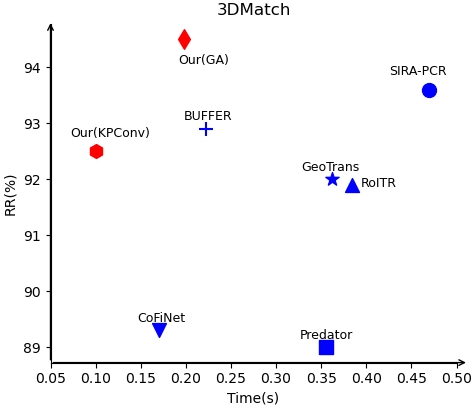}
        % \caption{Description for the first image.}
        % \label{fig:Mem_Time-3D}
    \end{minipage}%\hfill
    \begin{minipage}[r]{0.24\textwidth}
        \centering
        \includegraphics[width=\textwidth]{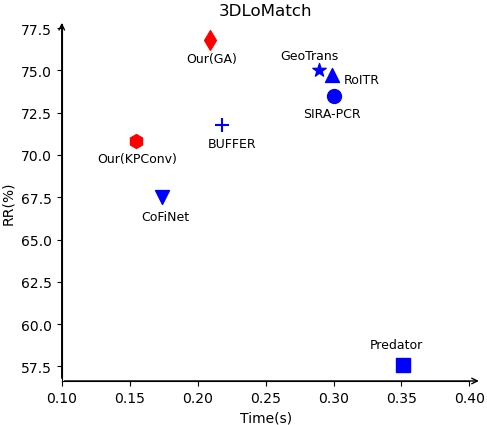}
        % \caption{Description for the second image.}
        % \label{fig:Mem_Time-3DLo}
    \end{minipage}
    % \begin{minipage}[r]{0.22\textwidth}
    % \makeatletter\def\@captype{subfigure}
    % \includegraphics[width=1\linewidth]{figs/iteration.pdf}
    % % \caption{}  % Iteration results
    % % \label{fig:iteration}
  % \end{minipage}
\caption{Comparison with existing methods on multiple datasets. Our method achieves SOTA registration accuracy (RR) with much lower time and GPU memory consumption on 3DMatch and 3DLoMatch datasets.}
\label{fig:Mem and Time on 3DLo}
\vspace{-2em}
\end{figure}

\begin{figure*}[t!]
  \centering
  \includegraphics[width=1.0\textwidth]{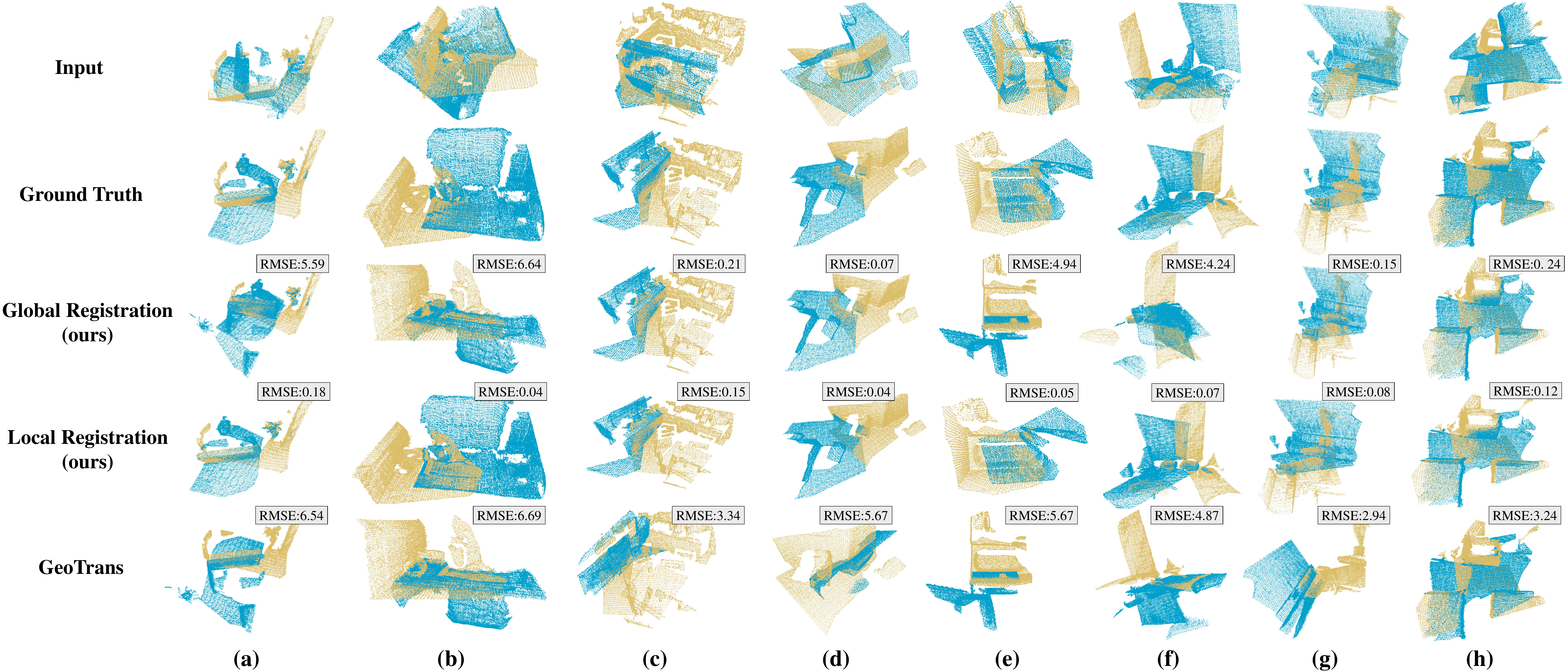}
  % \vspace{-1em}
  \caption{Visualization results of our network for 3DLoMatch (a-d) and 3DMatch (e-h) with global and local registration. Compared to GeoTrans~\cite{qin2022geometric}, our method has similar results in the global registration, and the local registration is further optimized to achieve superior results.}
\label{fig:result}
\vspace{-1em}
\end{figure*}

\paragraph{\textbf{Results.}} 
% \noindent \textit{Results.} 
Our method is compared with eight SOTA methods on registration results (RR), runtime (including data processing, model execution, pose estimation, and total time) and GPU memory usage on the 3DMatch and 3DLoMatch datasets, including CoFiNet, Predator, GeoTrans, REGTR, RoITR, BUFFER, RoReg, and SIRA-PCR. We conducted multiple comparative experiments based on different transformation estimator. Our method employs LGR during the registration process to obtain accurate pairing relationships, thereby eliminating the need for subsequent calculations using RANSAC. As shown in Table~\ref{tab:3d/3dlo} and Figure~\ref{fig:Mem and Time on 3DLo}, our method (GA for downsampling) achieves SOTA results on both datasets. Compared with methods SIRA-PCR and RoReg with similar results, our method leads the running time of the two data sets by $31\%$ and $26\%$. In order to prove the effectiveness of our proposed method, we also use the same KPConv-based sampling method as GeoTrans to conduct experiments. The results show that our method (KPConv for downsampling) achieves the fastest running time and second slightest GPU memory usage. The visual results are shown in Figure~\ref{fig:result}. Our method (GA) achieves lower RMSE errors during the global registration stage compared to GeoTrans. Additionally, scenes that initially fail in registration successfully align after undergoing local registration.

%-------------------- Table 1  Overall Performance ---------------------%
\begin{table*}[t]
\tiny
\caption{Registration results on 3DMatch and 3DLoMatch datasets. Our method achieves SOTA registration accuracy while using the lowest time and GPU memory consumption. The best and second-best results are marked in bold and underlined respectively to highlight the efficiency and accuracy advantages of our method with different sampling networks and estimators.}
\label{tab:3d/3dlo}
\resizebox{0.98\textwidth}{!}{
\centering
\renewcommand{\arraystretch}{1.3}
    \begin{tabular}{@{}c|c|c|clccccccc|lllllllll@{}}
    \toprule
    \multirow{3}{*}{Methods} & \multirow{3}{*}{Estimator} & \multirow{3}{*}{Sample} & \multicolumn{9}{c|}{3DMatch} & \multicolumn{9}{c}{3DLoMatch} \\ \cmidrule(l){4-21} 
    &  &  &\multicolumn{4}{c|}{Time (s)↓} & \multicolumn{1}{l|}{\textbf{Mem}} & \multicolumn{2}{c}{\textbf{RR}} & RTE & \multicolumn{1}{l|}{RRE} & \multicolumn{4}{c|}{Time (s)↓} & \multicolumn{1}{c|}{\textbf{Mem}} & \multicolumn{2}{c}{\textbf{RR}} & RTE & RRE \\
    & & & \textbf{Total} & Data & Model & \multicolumn{1}{l|}{Pose} & \multicolumn{1}{l|}{(GB)↓} & \multicolumn{2}{c}{(\%)↑} & (cm)↓ & \multicolumn{1}{c|}{(°)↓}  & \textbf{Total} & Data & Model & \multicolumn{1}{l|}{Pose} & \multicolumn{1}{l|}{(GB)↓}  & \multicolumn{2}{c}{(\%)↑} & \multicolumn{1}{c}{(cm)↓} & \multicolumn{1}{c}{(°)↓} \\ 
    \midrule
    
    Predator~\cite{huangpredator} & RANSAC-$50k$ & 1000 & 0.355 & 0.267 & 0.029 & \multicolumn{1}{c|}{0.059} & \multicolumn{1}{c|}{2.680} & \multicolumn{2}{c}{89.0} & 6.6 & 2.015 & 0.351 & 0.267 & 0.028 & \multicolumn{1}{l|}{0.056} & \multicolumn{1}{l|}{2.241} & \multicolumn{2}{c}{56.7} & \multicolumn{1}{c}{9.8} & 3.741  \\
    
    CoFiNet~\cite{yu2021cofinet} & RANSAC-$50k$ & 5000 & 0.170 & 0.030 & 0.091 & \multicolumn{1}{l|}{0.049} & \multicolumn{1}{c|}{3.472} & \multicolumn{2}{c}{89.3} & 6.4 & \multicolumn{1}{c|}{2.052} & 0.174 & 0.030 & 0.094 & \multicolumn{1}{l|}{0.050} & \multicolumn{1}{l|}{2.159} & \multicolumn{2}{c}{67.5} & \multicolumn{1}{c}{9.0} & 3.180 \\
    
    REGTR~\cite{yew2022regtr} & RANSAC-$50k$ & 5000 & 1.283 & 0.016 & 0.217 & \multicolumn{1}{c|}{1.050} & \multicolumn{1}{c|}{2.413} & \multicolumn{2}{c}{86.7} & 5.3 & \multicolumn{1}{c|}{1.774}  & 1.031 & 0.015 & 0.211 & \multicolumn{1}{l|}{0.785} & \multicolumn{1}{l|}{2.264} & \multicolumn{2}{c}{50.5} & \multicolumn{1}{c}{8.2} & 3.005  \\
    
    GeoTrans~\cite{qin2022geometric} & RANSAC-$50k$ & 5000 & 0.362 & 0.091 & 0.079 & \multicolumn{1}{c|}{0.192} & \multicolumn{1}{c|}{3.669} & \multicolumn{2}{c}{92.0} & 7.2 & \multicolumn{1}{c|}{2.130} & 0.289 & 0.088 & 0.080 & \multicolumn{1}{l|}{0.121} & \multicolumn{1}{l|}{3.677}& \multicolumn{2}{c}{75.0} & \multicolumn{1}{c}{12.6} & 3.558  \\
    
    RolTR~\cite{yu2023rotation} & RANSAC-$50k$ & 5000 & 0.384 & 0.022 & 0.214 & \multicolumn{1}{c|}{0.148} & \multicolumn{1}{c|}{2.507} & \multicolumn{2}{c}{91.9} & 5.3 & \multicolumn{1}{c|}{1.644} & 0.299 & 0.051 & 0.210 & \multicolumn{1}{l|}{0.067} & \multicolumn{1}{l|}{2.567}& \multicolumn{2}{c}{74.7} & \multicolumn{1}{c}{7.8} & 2.554  \\
    
    BUFFER~\cite{ao2023buffer} & RANSAC-$50k$ & 5000 & 0.222 & 0.033 & 0.186 & \multicolumn{1}{c|}{0.003} & \multicolumn{1}{c|}{4.976} & \multicolumn{2}{c}{92.9} & 6.0 & \multicolumn{1}{c|}{1.863}  & 0.218 & 0.033 & 0.182 &\multicolumn{1}{l|}{0.003 } & \multicolumn{1}{l|}{4.976}& \multicolumn{2}{c}{71.8} & \multicolumn{1}{c}{10.1} & 3.018  \\

    RoReg~\cite{wang2023roreg} & RANSAC-$50k$ & 5000 & 3.490 & 0.087 & 0.050 & \multicolumn{1}{c|}{2.353} & \multicolumn{1}{c|}{9.663} & \multicolumn{2}{c}{93.2} & 6.3 & \multicolumn{1}{c|}{1.840}  & 3.490 & 0.087 & 0.050 & \multicolumn{1}{c|}{2.353} & \multicolumn{1}{c|}{9.662} & \multicolumn{2}{c}{71.2} & \multicolumn{1}{c}{9.3} & 3.090  \\

    SIRA-PCR~\cite{chen2023sira} & RANSAC-$50k$ & 5000 & 0.469 & 0.141 & 0.144 & \multicolumn{1}{c|}{0.184} & \multicolumn{1}{c|}{2.674} & \multicolumn{2}{c}{93.6} & 6.3 & \multicolumn{1}{c|}{1.862}  & 0.300 & 0.140 & 0.139 &\multicolumn{1}{l|}{0.021} & \multicolumn{1}{l|}{2.627}& \multicolumn{2}{c}{73.5} & \multicolumn{1}{c}{8.5} & 2.846  \\ 
    
    REGTR~\cite{yew2022regtr} & weighted SVD & 250 & 0.229 & 0.013 & 0.215 & \multicolumn{1}{c|}{0.001} & \multicolumn{1}{c|}{2.413} & \multicolumn{2}{c}{92.0} & 4.9 & \multicolumn{1}{c|}{1.562} & 0.221 & 0.014 & 0.206 & \multicolumn{1}{l|}{0.001} & \multicolumn{1}{l|}{2.328} & \multicolumn{2}{c}{64.8} & \multicolumn{1}{c}{7.8} & 2.767  \\
    
    GeoTrans~\cite{qin2022geometric} & weighted SVD & 250 & 0.172 & 0.091 & 0.079 & \multicolumn{1}{c|}{0.002} & \multicolumn{1}{c|}{3.669} & \multicolumn{2}{c}{86.5} & 6.7 & \multicolumn{1}{c|}{2.043} & 0.170 & 0.088 & 0.080 & \multicolumn{1}{l|}{0.002} & \multicolumn{1}{l|}{3.677} & \multicolumn{2}{c}{59.9} & \multicolumn{1}{c}{10.2} & 3.709  \\
    
    BUFFER~\cite{ao2023buffer} & weighted SVD & 250 & 0.224 & 0.033 & 0.189 & \multicolumn{1}{c|}{0.002} & \multicolumn{1}{c|}{4.978} & \multicolumn{2}{c}{92.1} & 6.1 & \multicolumn{1}{c|}{1.881}  & 0.214 & 0.033 & 0.189 & \multicolumn{1}{l|}{0.002} & \multicolumn{1}{l|}{4.774}& \multicolumn{2}{c}{70.0} & \multicolumn{1}{c}{10.0} & 3.026  \\

    RolTR~\cite{yu2023rotation} & LGR & all & 0.269 & 0.048 & 0.214 & \multicolumn{1}{c|}{0.007} & \multicolumn{1}{c|}{2.641} & \multicolumn{2}{c}{90.9} & 5.2 & \multicolumn{1}{c|}{1.640} & 0.266 & 0.050 & 0.210 & \multicolumn{1}{l|}{0.006} & \multicolumn{1}{l|}{2.617} & \multicolumn{2}{c}{73.9} & \multicolumn{1}{c}{7.4} & 2.499 \\
    
    GeoTrans~\cite{qin2022geometric} & LGR & all & 0.178 & 0.091 & 0.079 & \multicolumn{1}{c|}{0.008} & \multicolumn{1}{c|}{3.669} & \multicolumn{2}{c}{91.5} & 6.3 & \multicolumn{1}{c|}{1.808}& 0.175 & 0.088 & 0.080 & \multicolumn{1}{l|}{0.007} & \multicolumn{1}{l|}{3.677} & \multicolumn{2}{c}{74.0} & \multicolumn{1}{c}{8.9} & 2.936 \\

    SIRA-PCR~\cite{chen2023sira} & LGR & all & 0.290 & 0.141 & 0.144 & \multicolumn{1}{c|}{0.005} & \multicolumn{1}{c|}{2.674} & \multicolumn{2}{c}{94.1} & 5.1 & \multicolumn{1}{c|}{1.539}  & 0.284 & 0.140 & 0.139 &\multicolumn{1}{l|}{ 0.005} & \multicolumn{1}{l|}{2.627}& \multicolumn{2}{c}{\underline{76.6}} & \multicolumn{1}{c}{7.2} & 2.388  \\ 
    
    \midrule
    
    Ours (KPConv) & weighted SVD & 250 & \textbf{0.097} & 0.048 & 0.047 & \multicolumn{1}{c|}{0.002} & \multicolumn{1}{c|}{\textbf{1.547}} & \multicolumn{2}{c}{{92.5}} & 6.4 & \multicolumn{1}{c|}{1.917} & \textbf{0.152} & 0.052 & 0.098 & \multicolumn{1}{l|}{0.002} & \multicolumn{1}{l|}{\textbf{1.455}}& \multicolumn{2}{c}{70.3} & \multicolumn{1}{c}{9.4} & 3.254  \\ 
    
    Ours (GA) & weighted SVD & 250 & 0.195 & 0.012 & 0.179 & \multicolumn{1}{c|}{0.002} & \multicolumn{1}{c|}{\underline{1.883}} & \multicolumn{2}{c}{{\underline{94.4}}} & 6.4 & \multicolumn{1}{c|}{1.755} & 0.205 & 0.013 & 0.190 & \multicolumn{1}{l|}{0.002} & \multicolumn{1}{l|}{2.234}& \multicolumn{2}{c}{74.4} & \multicolumn{1}{c}{8.7} & 3.002  \\     
    
    Ours (KPConv) & LGR & all & \underline{0.100} & 0.048 & 0.047 & \multicolumn{1}{c|}{0.005} & \multicolumn{1}{c|}{\textbf{1.547}} & \multicolumn{2}{c}{92.5} & 6.4 & \multicolumn{1}{c|}{1.897} & \underline{0.155} & 0.052 & 0.098 & \multicolumn{1}{l|}{0.005} & \multicolumn{1}{l|}{\underline{1.456}}& \multicolumn{2}{c}{70.8} & \multicolumn{1}{c}{9.4} & 3.254 \\ 

    Ours (GA) & LGR & all & 0.198 & 0.012 & 0.179 & \multicolumn{1}{c|}{0.005} & \multicolumn{1}{c|}{\underline{1.883}} & \multicolumn{2}{c}{{\textbf{94.5}}} & 6.4 & \multicolumn{1}{c|}{1.772} & 0.209 & 0.013 & 0.191 & \multicolumn{1}{l|}{0.005} & \multicolumn{1}{l|}{2.234}& \multicolumn{2}{c}{\textbf{76.8}} & \multicolumn{1}{c}{9.1} & 2.989  \\ \bottomrule
    \end{tabular}
    }
    \vspace{-1em}
\end{table*}

%-------------------------------------------%
\begin{figure}[t]%{0.5\textwidth}
    \includegraphics[width=1.05\linewidth]{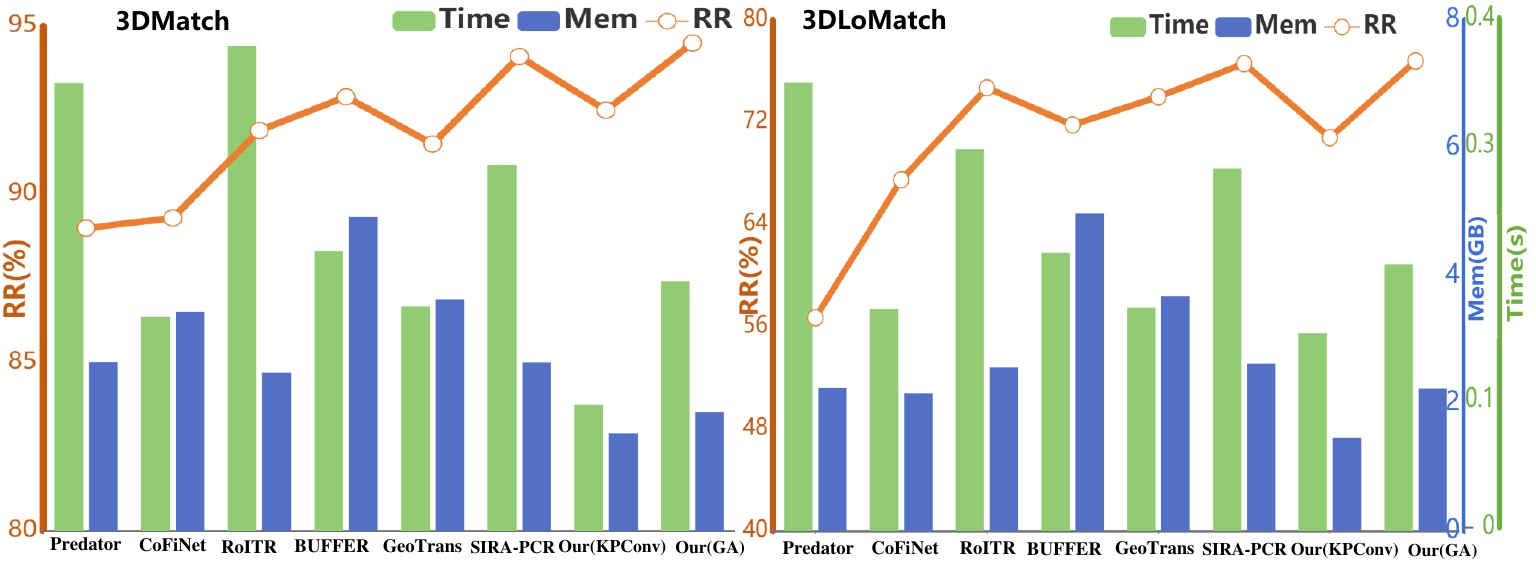}
\end{figure}

%---------------- Figure Mem Time 3D-------------------%
%----------------------------------------%

\subsection{Outdoor Benchmark: KITTI odometry}
\paragraph{\textbf{Dataset and Metrics.}} The KITTI odometry~\cite{geiger2012we} dataset comprises $11$ LiDAR-scanned outdoor driving sequences. 
% In line with previous work~\cite{bai2021pointdsc, yang2022one}, we partition the dataset into training sequences $0-5$, validation sequences $6-7$, and testing sequences $8-10$. 
Same as 3DMatch, our method is assessed using three key metrics: RRE, RTE and RR which denotes the proportion of point cloud pairs meeting specific criteria (i.e., RRE \textless $5^\circ$ and RTE \textless $2m$).
\label{sec:kitty}

\paragraph{\textbf{Results.}} We compared our method (KPConv for downsampling) with FCG, RegFormer, as well as four methods Predator, CoFiNet, GeoTrans, BUFFER and  that use KPConv for downsampling on the KITTI dataset. Under the condition of using the same sampling method, our approach achieves SOTA performance on RR while surpassing the second fastest method by over $33\%$ in terms of runtime, and it attains the second smallest result in GPU memory consumption, using less than $2.4 GB$. Although FCGF consumes fewer GPU resources, it spends more time on pose estimation, leading to significantly longer runtime compared to other methods. Our network achieves comparable performance in RTE and RRE even though our sampling is sparser, and for the most critical registration metric RR, we achieve the best performance.

%------------------------ Table 2  KITTI ------------------------%
\begin{table}
% \tiny
\caption{Results on KITTI odometry dataset. The best and second-best results are marked in bold and underline, respectively. Our method achieved the best RR and Time.}
\label{tab:kitti}
\small
  \centering
  \renewcommand{\arraystretch}{1.3}
  \fontsize{6}{6}\selectfont
  \resizebox{0.48\textwidth}{!}{
      \begin{tabular}{c|c c|c c c}
        \toprule
         \multirow{2}{*}{Method} & \textbf{Time} & \textbf{Mem} & \textbf{RR} & RTE & RRE \\
                                 & (s)↓ &(GB)↓ &(\%)↑ &(cm)↓ &(°)↓ \\
        \midrule
        FCGF~\cite{choy2019fully} & 7.454 & \textbf{1.788} & 98.9 & 6.0 & 0.39 \\
        RegFormer~\cite{liu2023regformer} & \underline{0.205} & \underline{2.242} & \underline{99.5} & 9.7 & 0.22 \\
        \midrule
        Predator~\cite{huangpredator} & 0.703 & 4.634 & \textbf{99.8} & 6.8 & 0.26 \\
        CoFiNet~\cite{yu2021cofinet} & 0.277 & 3.055 & \textbf{99.8} & 8.2 & 0.41\\
        GeoTrans~\cite{qin2022geometric} & {0.251} & 3.666  & \textbf{99.8} & 6.2 & 0.23 \\
        BUFFER~\cite{ao2023buffer} & 0.450 & 5.682 & 97.5  & 6.3 & 0.23\\
        Our(KPConv) & \textbf{0.168} & 2.304  & \textbf{99.8} & 7.8 & 0.34 \\
        \bottomrule
      \end{tabular}
  }
  % \vspace{-2em}
\end{table}
%------------------------------------------------------------%

%%%%%%%%%%%%%%%%%%%%%%%%%%%%%%%%%%%%%%%%%%
\subsection{Ablation Study}
\label{sec:ablation}
We show the ablation studies of our method (GA) on the 3DLoMatch dataset.

\paragraph{\textbf{Global and Local Encoder.}} We test not using two identical encoders in global registration and local registration. For the global registration stage, the data of global sampling points pays more attention to learning the global representation. After multiple downsampling, the greater sparsity between global point clouds leads to larger feature differences between them. For the sampling points used in local registration, more emphasis is placed on learning local information within the neighborhood. Therefore, as shown in Table~\ref{tab:ablation} (a), our method uses two encoders to facilitate enhanced feature learning at different stages, contributing to improved results.

%-------------------------- Table 3 Ablation Study -------------------------%
\begin{table}
\tiny
\caption{Ablation study on 3DLoMatch dataset. The best results are marked in bold.}
\label{tab:ablation}
\tiny% \small
  \centering
  % \fontsize{8}{10}\selectfont
  \resizebox{0.45\textwidth}{!}{
      \begin{tabular}{p{1.2cm} p{2.4cm}|p{0.6cm}<{\centering} p{0.5cm}<{\centering} p{0.4cm}<{\centering} p{0.5cm}<{\centering}}
        \toprule
        \multirow{2}{*}{Module} & \multirow{2}{*}{Model} & \multicolumn{4}{c}{3DLoMatch}\\
        & & Time & \textbf{RR} & RRE & RTE \\
        % \hline
        \midrule
        \multirow{2}{*}{a. Encoder} & 1. unique encoder & \textbf{0.200} & 74.4 & 2.990& 9.1\\
                                    & 2. Ours & 0.209  & \textbf{76.8} & 2.989 & 9.1 \\
        \midrule
        \multirow{2}{*}{b. SC Classifier} & 1. w/o classifier & 0.234& 72.7& 3.032& 9.2\\
                                       & 2. Ours & \textbf{0.209} & \textbf{76.8} & 2.989 & 9.1\\
        \midrule
        \multirow{2}{*}{c. Node} & 1. random nodes & 0.202 & 74.4 & 2.969 & 9.1\\
                                 & 2. average center & \textbf{0.195} & 73.4 & 3.008 & 9.2\\
                                 & 3. DBSCAN(Ours) & 0.209 & \textbf{76.8}& 2.989 & 9.1\\
        \midrule
        \multirow{4}{*}{d. Iteration} & 1. iteration=0 & \textbf{0.183}& 73.6 & 3.010& 9.3\\
                                      & 2. iteration=1 & 0.196 & 74.4 & 2.975 & 9.2\\
                                      & 3. iteration=2 & 0.201& 76.2 & 3.045 & 9.2\\
                                      & 4. iteration=3 & 0.208 & 76.6 & 3.056 & 9.2\\
                                      & 5. iteration=4(Ours) & 0.209 & \textbf{76.8} & 2.989 & 9.1\\
        \bottomrule
      \end{tabular}
  }
  % \vspace{-1em}
\end{table}
%------------------------------------------------------------------%

%------------------- Figure Visualization ---------------------%
%--------------------------------------------------------------%

\paragraph{\textbf{SC Classifier.}} We implement the early-exit mechanism for the iterative registration process by setting a numerical threshold for first-order spatial consistency (\(SC\)).

\begin{figure}
\tiny
  \centering
  \begin{minipage}[c]{0.2\textwidth}
    \centering
    \makeatletter\def\@captype{subfigure}
    \fontsize{8}{10}\selectfont
    \resizebox{0.99\textwidth}{!}{
        \begin{tabular}{p{0.6cm}<{\centering} p{0.2cm}<{\centering} p{0.4cm}<{\centering} |p{0.3cm}<{\centering} p{0.4cm}<{\centering}}
            \toprule
            Metrics & \(N_{g}\) & \(N_{l}^{0}\) & RR & Time\\
            % \hline
            \midrule
            {\(IR\)} & 0.30 & -- & 73.4 & \textbf{0.190}\\
            {\(NC\)} & 150 & -- & 73.5 & 0.201\\
            {\(SC^{2}\)} & 150 & -- & 72.7 & 0.209\\
            {\(SC\)} & 150 & -- & \textbf{73.7} & 0.200\\
            \midrule
            \multirow{3}{*}{\(SC\)} & $100$ & 10 & 74.0 & \textbf{0.194}\\
                                    & $150$ & 15 & \textbf{74.4} & 0.196 \\
                                    & $200$ & 20 & \textbf{74.4} & 0.208 \\
            \bottomrule
        \end{tabular}
    }
    % \caption{}  % Metrics and Thresholds
    % \label{tab:metrics}
    \end{minipage}
    \hspace{0.02\textwidth}
   \begin{minipage}[c]{0.21\textwidth}
        \makeatletter\def\@captype{subfigure}
        \includegraphics[width=1.0\linewidth]{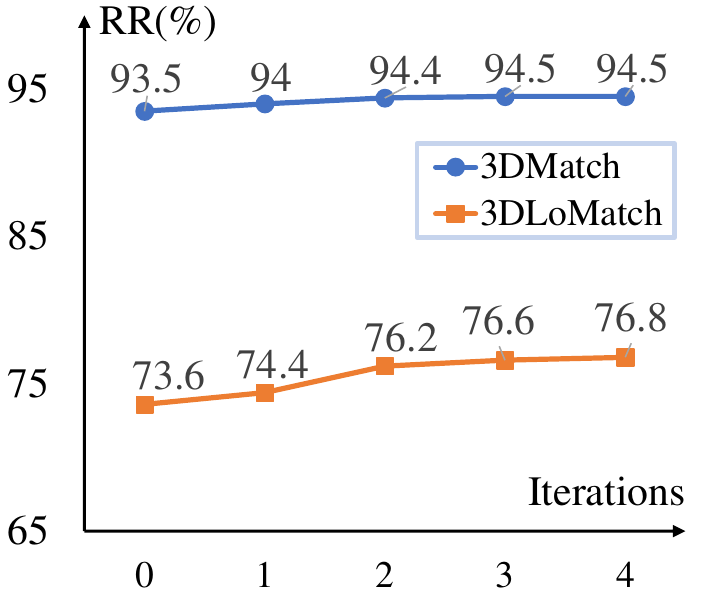}
        % \caption{}  % Iteration results
        % \label{fig:iteration}
   \end{minipage}
\makeatletter\def\@captype{table}
% \caption{Comparison of different metrics in global registration and different \(SC\) thresholds (\(N_{l}^{i}\)) during local registration iterations.} 
% \label{tab:metrics}
\caption{ Left: Comparison of different metrics and different \(SC\) thresholds (\(N_{g}\)). Right: The iteration results on 3DMatch/3DLoMatch demonstrates the effectiveness and stability of our iterations. } 
\label{fig:iteration}
\makeatletter\def\@captype{figure}
\vspace{-1em}
\end{figure}

\begin{enumerate}
\setlength{\itemsep}{0pt}
\setlength{\parsep}{0pt}
\setlength{\parskip}{0pt}
    \item \textbf{Without Classifier.} We test the results without adding the classifier. As can be seen from Table~\ref{tab:ablation} (b), the classifier achieves the purpose of reducing computing resources because it can effectively filter out a large number of simple registration scenarios.
    \item \textbf{Different Metrics.} We assess the performance of model after the first local registration using different metrics as the classifier in Table~\ref{fig:iteration}, including inlier ratio (\(IR\)), normal consistency (\(NC\)), second-order spatial consistency (\(SC^{2}\)) and spatial consistency \(SC\). Based on the running time and RR of each metric, \(SC\) performs the best.
    \item \textbf{Different Thresholds.} We experimented with different global \(SC\) thresholds \(N_{g}\) and local \(SC\) comparison threshold \(N_{l}^{i}\) and the results are shown in Table~\ref{fig:iteration}. For the local registration stage, We initially set the threshold for the first local \(SC\) comparison threshold \(N_{l}^{1}\) to $0.1$ of the global \(SC\) threshold and set the comparison threshold change difference to \(\delta N_{l}^{i}\) (\(\delta N_{l}^{i}\)=10 by default), which represents the number of \(N_{l}^{i+1}\) is increased compared to \(N_{l}^{i}\) each time (i.e., \(N_{l}^{i}\)=[$15$, $25$, $35$...]). 
\end{enumerate}

%---------------- Table 4  左Table + 右 Figure -------------------%
%--------------------------------------------------------%

\paragraph{\textbf{Refined Nodes.}} Before performing local registration, we utilize the Refined Nodes module to select new nodes for registration based on the global matched points obtained from the last registration. We compare the DBSCAN clustering algorithm with random sampling and global centers methods, and the results in Table~\ref{tab:ablation} (c) demonstrate that while our method may consume more time, it significantly improves the registration results.

\paragraph{\textbf{Iteration.}} In order to prove the effect of iteration of our method, we compare the method global registration Table~\ref{tab:ablation} (d.1) without iteration with the method using local registration multiple times Table~\ref{tab:ablation} (d.2-4) for iteration. Additionally, we plot the iterative results in the line chart shown in Table~\ref{fig:iteration} on the 3DMatch and 3DLomatch datasets. As the number of iterations increases, the registration results gradually increase starting from 73.6. After the fourth iteration, our method reached the maximum value of 76.8 on the 3DLoMatch dataset. Similarly, we reached the maximum value of 94.5 on the 3DMatch dataset after three iterations and remained unchanged in subsequent iterations. However, during multiple iterations, the RRE continues to increase. This is because, in the process of dynamically finding the registration region, our model continues to reduce the number and region of registration points, which will affect the finer-grained accuracy in some scenes. But it still performs well on the more important metric RR.

\section{Conclusion}
\label{sec:conclusion}
We innovatively introduce an efficient Point Cloud Registration via Dynamic Iteration. 
Our Deeper Sampling reduces significantly the computational complexity of the network, and our Refined Nodes module discovers the inlier concentration region among the paired points for the next registration. The proposed iterative approach dynamically narrows down the registration region to eliminate interference from outliers, and then the iterations are terminated once the registration result meets our SC classifier condition threshold. Extensive experiments demonstrate that our \mymodel achieves the best RR result on both 3DMatch and 3DLoMatch datasets while having the advantage of computational efficiency over existing methods.

\bigskip

\newpage
\bibliography{main}

\appendix

% \section{Related Work}
% supplementary
\title{Supplementary Materials: A Dynamic Network for Efficient Point Cloud Registration}
% \author{Anonymous Authors}
\author{}
\maketitle

\setcounter{table}{0}
\setcounter{figure}{0}
\setcounter{equation}{0}

\section{Supplement of Related Work}

\paragraph{Correspondence-based Methods}
DGR~\cite{choy2020deep} and 3DRegNet~\cite{pais20203dregnet} consider the inlier/outlier determination as a classification problem. MAC~\cite{zhang20233d} developed a compatibility graph to render the affinity relationship between correspondences and construct the maximal cliques to represent consensus sets. 

\paragraph{Deep-learned Feature Descriptors} 
PointNetLK~\cite{aoki2019pointnetlk} and PCRNet~\cite{sarode2019pcrnet} integrate PointNet~\cite{qi2017pointnet} into the point registration task and other deep learning methods~\cite{deng2018ppfnet, yew2020rpm} improve feature extraction module to enhances the capability of feature descriptors. 
CoFiNet~\cite{yu2021cofinet} leverages a two-stage method, using down-sampling and similarity matrices for rough matching and up-sampling the paired point neighborhood to obtain fine matching. 
Additionally, RoITR~\cite{yu2023rotation} addresses rotation-invariant problems by introducing novel local attention mechanisms and global rotation-invariant transformers. 
BUFFER~\cite{ao2023buffer} attempts to improve the computation efficiency, but its performance is mainly reflected in its generalization ability. 

\paragraph{Dynamic Neural Networks} 
Compared to static networks, dynamic networks can adapt parameters or structures to different inputs, leading to notable advantages in terms of accuracy, computational efficiency, adaptiveness, etc~\cite{han2021dynamic}. 
In recent years, many advanced dynamic network works have been proposed and mainly divided into three different aspects: 1) sample-wise dynamic networks~\cite{yang2019condconv} are designed to adjust network architectures and parameters to reduce computation; 2) temporal-wise dynamic networks~\cite{meng2020adafuse} treat video data as a sequence of images to save computation along temporal dimension; 
3) spatial-wise dynamic networks~\cite{zhou2016learning} exploit spatial redundancy to improve efficiency.

\section{Network Architecture Details}
% We first introduce the two sampling methods KPConv in Sec.~\ref{KPConv} and Global aggregation in Sec.~\ref{GA} used in the paper, and then introduce the encoder module in Sec.~\ref{encoder} used in the global and local registration process.

We first introduce the details of the two sampling networks we used, KPConv and Global aggregation, and then introduce the encoder module used in our global and local registration stages.

\subsection{KPconv Network}
\label{KPConv}
We use the KPConv~\cite{thomas2019kpconv} network as our backbone for multi-level downsampling and feature extraction. The specific structure is the same as GeoTrans~\cite{qin2022geometric}. First, the input point cloud is sampled using grid downsampling with a voxel size of $2.5cm$ on 3DMatch/3DLoMatch and $30cm$ on KITTI. With each subsequent downsampling operation, the voxel size is doubled. We use Deeper Sampling to obtain sparser point clouds, respectively, 5-stage for 3DMatch and 6-stage for KITTI. The numbers of calibrated neighbors used for downsampling are \{$37$, $36$, $36$, $38$, $36$\} for 3DMatch and \{$65$, $63$, $70$, $74$, $69$, $58$\} for KITTI. Other settings are consistent with GeoTrans~\cite{qin2022geometric}.

\subsection{Global Aggregation}
\label{GA}

We use the Global Aggregation module proposed in RoITr~\cite{yu2023rotation} as our downsampling network for multi-level point cloud sampling and initial feature extraction. Different from the RoITr~\cite{yu2023rotation}, the Global Aggregation we use not only includes the global transformer module, but also includes the complete encoder-decoder structure. The encoder part of the network inputs coordinates, normal vectors and initial features as a pair of triples \(\mathcal{P}=(P, N, X)\) and \(\mathcal{Q}=(Q, M, Y)\) to downsamples the origin points to coarser nodes triples \(\mathcal{P^{'}}\) and \(\mathcal{Q^{'}}\). Then, the global transformer aggregates the geometric cues to enhance the features of coarse nodes by the consecutive cross-frame context aggregation. And the decoder module of the network upsample the refined nodes to the original input size through multi-stage upsampling. We also use Deeper Sampling in the network to reduce the number of global registration, while not performing complete upsampling. Here, we set the number of downsampling layers to 5 times, and the number of upsampling layers is 2. Other settings are consistent with RoITr~\cite{yu2023rotation}. %Finally, the upsampled point cloud and its features are used in the local registration process.

\subsection{Geometric Transformer Encoder}
\label{encoder}
We use the Geometric Transformer in GeoTrans~\cite{qin2022geometric} as our encoder in the global registration stage and local registration stage. Firstly, the encoder receives the features from KPConv and projects them to $d=256$ for 3DMatch/3DLoMatch and $d=128$ dimensions for KITTI. For the local encoder, since the features come from the third level of KPConv sampling, the dimensions are from $512$ to $256$ and $512$ to $128$ when performing feature projection. Then after three alternations of `self-attention and cross-attention', the features are obtained and used for pairing point sets through feed-forward networks (FFN).
%%%%%%%%%%%%%%%%%%%%%%%%%%%%%%%%%%%%%%%%%%%%%%%%%%%%%%%%%%%%%%%%%%%%5

\section{Implementation Details and Parameters}
Our proposed method was implemented by PyTorch. To ensure fairness, we employed the official code and pre-trained models provided by the baseline methods for comparison. All experiments were performed on the same PC with an Intel Core i7-12700K CPU and a single Nvidia RTX 3090 with $24G$ memory. We train the model for $40$ epochs on both 3DMatch~\cite{zeng20173dmatch} and KITTI~\cite{geiger2012we} with the batchsize of $1$. We use an Adam optimizer with an initial learning rate of $1e-4$, which is exponentially decayed by $0.05$ every epoch for 3DMatch and every $4$ epoch for KITTI. And we set the numbers of downsampling layers as $5$ for the 3DMatch dataset and $6$ for the KITTI dataset. The numbers of points in the patch are set to $32$ and $64$, respectively. Other parameters remain consistent with GeoTrans~\cite{qin2022geometric}.
%\subsection{Refined Nodes Module Parameters}
\paragraph{Refined Nodes Parameters}
For our refined nodes module, we use the density-based clustering method (DBSCAN) ~\cite{ester1996density} combined with the pairing similarity to cluster paired point clouds and form new registration nodes. We set the minimum number of samples for each category to $3$ for 3DMatch~\cite{zeng20173dmatch} and $5$ for 3DLoMatch and KITTI~\cite{geiger2012we}  dataset. The maximum clustering radius are set to gradually increase with the iteration of the registration process. For 3DMatch, the maximum clustering radius is \{$0.125$, $0.250$, $0.375$, $0.500$\}, and for 3DLoMatch, the maximum clustering radius is \{$0.25$, $0.30$, $0.35$, $0.40$\}, while for KITTI it is a fixed value of $30$.
%\subsection{Classifier Parameters}
\paragraph{ SC Classifier Parameters}
In the inference phase, for the classifier threshold, we set the global \(SC\) threshold to $200$ for the 3DMatch dataset, and the local \(SC\) comparison threshold (limiting the reduction in the number of pairs) to \{$20$, $0$, $-20$, $-40$\}. Similarly, for 3DLoMatch, we set the global threshold to $150$ and the iteration threshold to \{$15$, $25$, $35$, $45$\}, and for KITTI, we set the global threshold to $30$ and the iteration threshold to \{$10$, $0$, $-10$\}.

\section{Loss Function}
\subsection{Coarse Matching Loss}

We use the overlap-aware circle loss to supervise the point-wise feature descriptors and focus the model on matches with high overlap following GeoTrans~\cite{qin2022geometric}. The overall coarse matching loss is $\mathcal L_{c} = (\mathcal L_{c}^{Q} + \mathcal L_{c}^{P}) / 2 $. Consider a pair of overlapping point clouds $P$ and $Q$ aligned by ground truth transformation. We choose the points $p_{i} \in P$ that have at least one correspondence in $Q$ to form a set of anchor patches, denoted as $\mathcal{A}$. For each anchor patch $\mathcal{G}_{i}^{\mathcal{P}} \in \mathcal{A}$, we consider the paired patches are positive if they share a minimum overlap of $10\%$, and negative if they have no overlap. We identify the set of its positive patches in $Q$ as $\varepsilon_{p}^{i}$, and the set of its negative patches as $\varepsilon_{n}^{i}$.
The coarse matching loss on $P$ is defined as:
\begin{equation} 
% \begin{split}
\small
\mathcal L_{c}^{P}\!=\!\frac{1}{|\mathcal A|} \! \sum_{\mathcal G_{i}^{\mathcal P} \in \mathcal A}log[1+\!\!\sum_{\mathcal G_{j}^{\mathcal{Q}}\in\varepsilon_{p}^{i}}\! e^{\lambda_{i}^{j}\beta_{p}^{i,j}(d_{i}-\Delta_{p})} \! \cdot \!\! \sum_{\mathcal G_{k}^{\mathcal{Q}}\in\varepsilon_{n}^{i}}\!\!e^{\beta_{n}^{i,k}(\Delta_{n}-d_{i}^{k})}]
% \end{split}
\end{equation}
where $d_{i}^{j}=||f_{p_{i}}-f_{q_{j}}||_{2}$ is the feature distance, $\lambda_{i}^{j}= (o_{i}^{j})^{\frac{1}{2}}$ and $o_{i}^{j}$ represents the overlap ratio between $\mathcal G_{i}^{\mathcal P}$ and $\mathcal G_{j}^{\mathcal Q}$. $\beta_{p}^{i,j}$ and $\beta_{n}^{i,k}$ are the positive and negative weights respectively, $\Delta_{n}$ and $\Delta_{p}$ are the margin hyper-parameters which are set to $0.1$ and $1.4$. The same goes for the loss $\mathcal L_{c}^{Q}$ on $Q$.

\subsection{Point Matching Loss}
During training, we randomly sample $N_{g}$ ground-truth node correspondences {$\hat{C_{i}^{*}}$} instead of using the predicted ones. We employ a negative log-likelihood loss on the assignment matrix $\Bar{C}_{i}$ for each sparse correspondences. The point matching loss is calculated by taking the average of the individual losses across all points correspondences:
$\mathcal{L}_{p} = \frac{1}{N_{g}} {\textstyle \sum_{i=1}^{N_{g}}}\mathcal{L}_{p,i}$. For each correspondence, a set of paired point $\mathcal{M}_i$ is extracted with a matching radius $\tau$. The unmatched points in the two patches are represented by $\mathcal{I}_{i}$ and $\mathcal{J}_{i}$. The individual point matching loss for correspondence $\hat{C_{i}^{*}}$ is computed as:
\begin{equation}
% \begin{split}
\mathcal{L}_{p,i} \!=\! -\!\!\sum_{(x,y)\in \mathcal{M}_{i}}\!\!log\bar{C}_{x,y}^{i} \!-\! \sum_{x\in \mathcal{I}_{i}} \!log\bar{C}_{x,m_{i}+1}^{i} \!-\! \sum_{y\in \mathcal{J}_{i}} \!log\bar{C}_{n_{i}+1, y}^{i}
% \end{split}
\end{equation}

\section{Metrics}
\label{Metrics supp}
We use the same metrics for 3DMatch/3DLoMatch and KITTI datasets, namely RTE, RRE, RR respectively.
% \vspace{-0.3cm}
\subsection{Relative Translation Error (RTE)} RTE specifically focuses on the translation component of the registration error. It measures the Euclidean distance between the true translation and the estimated translation achieved during the registration process. RTE provides insight into how well the registration algorithm performs in terms of aligning the point clouds along the translation axes.
\begin{equation} 
RTE = || t_{est} - t_{gt} ||
\end{equation}
where $t_{est}$ and $t_{gt}$ denote the estimated translation vector and ground truth translation vector, respectively. $||\cdot||$ represents the Euclidean norm
% \vspace{-0.5cm}
\subsection{Relative Rotation Error (RRE)}
RRE measures the angular difference between the true rotation and the estimated rotation after point cloud registration. It is a measure of how well the rotational alignment has been achieved.
\begin{equation} 
RRE = \arccos( \frac{trace(R^{T}_{est} R_{gt})-1}{2}) 
\end{equation}
where $R_{est}$ and $R_{gt}$ denote the estimated rotation matrix and ground truth rotation matrix, respectively. $arccos(\cdot)$ is the inverse cosine function, $trace(~\cdot)$ represents the trace of the matrix.
% \vspace{-0.5cm}

%------------------------------ Table Ablation Study -------------------------------------%
\begin{table}
\caption{Ablation study on 3DMatch dataset. The best results are marked in bold.}
\label{tab:supp_ablation}
\small
  \centering
  \fontsize{8}{10}\selectfont
  \begin{tabular}{p{1.2cm} p{2.4cm}|p{0.6cm}<{\centering} p{0.5cm}<{\centering} p{0.4cm}<{\centering} p{0.5cm}<{\centering}}
    \toprule
    \multirow{2}{*}{Module} & \multirow{2}{*}{Model} & \multicolumn{4}{c}{3DMatch}\\
    & & Time & RR & RRE & RTE \\
    % \hline
    \midrule
    \multirow{2}{*}{a. Encoder} & 1. unique encoder & \textbf{0.193} & 93.5 & 1.769 & 6.4\\
                                & 2. Ours & 0.198 & \textbf{94.5} & 1.772 & 6.4\\
    \midrule
    \multirow{2}{*}{b. SC Classifier} & 1. w/o classifier & 0.302 & 90.4 & 2.076 & 7.0\\
                                   & 2. Ours & \textbf{0.198} & \textbf{94.5} & 1.772 & 6.4\\
    \midrule
    \multirow{2}{*}{c. Node} & 1. random nodes & 0.192 & 93.6 & 1.773 & 6.4\\
                             & 2. average center & \textbf{0.190} & 93.5 & 1.757 & 6.3\\
                             & 3. DBSCAN (Ours) & 0.198 & \textbf{94.5} & 1.772 & 6.4\\
    \midrule
    \multirow{4}{*}{d. Iteration} & 1. iteration=0 & \textbf{0.187}& 93.5 & 1.768 & 6.4\\
                                  & 2. iteration=1 & 0.191 & 94.0 & 1.771 & 6.3\\
                                  & 3. iteration=2 & 0.196 & 94.4 & 1.771 & 6.2\\
                                  & 4. iteration=3 (Ours) & 0.198 & \textbf{94.5} & 1.772 & 6.3\\
                                  & 5. iteration=4 & 0.201 & \textbf{94.5} & 1.772 & 6.3\\
    \bottomrule
  \end{tabular}
\end{table}

% \section{Additional Qualitative Results}
% \label{supp_quality supp}

\subsection{Registration Recall (RR)}
Registration Recall is a measure of the accuracy of point cloud registration, specifically focusing on the overall transformation quality, considering both rotation and translation. 

% \begin{figure*}[t]%{0.5\textwidth}
%     \begin{minipage}{\textwidth}
%     \centering
%     \includegraphics[width=0.89\textwidth]{figs/result-kitti-supplement.pdf}
%     % \captionsetup{justification=justified, singlelinecheck=false}
%     \captionof{figure}{Registration results on KITTI. Our method achieves higher accuracy after local registration and achieves good performance similar to ground truth.}
%     \label{fig:result kitti}
%     \end{minipage}
% \end{figure*}

\begin{figure*}[!t]
    \centering
    \includegraphics[width=0.99\textwidth]{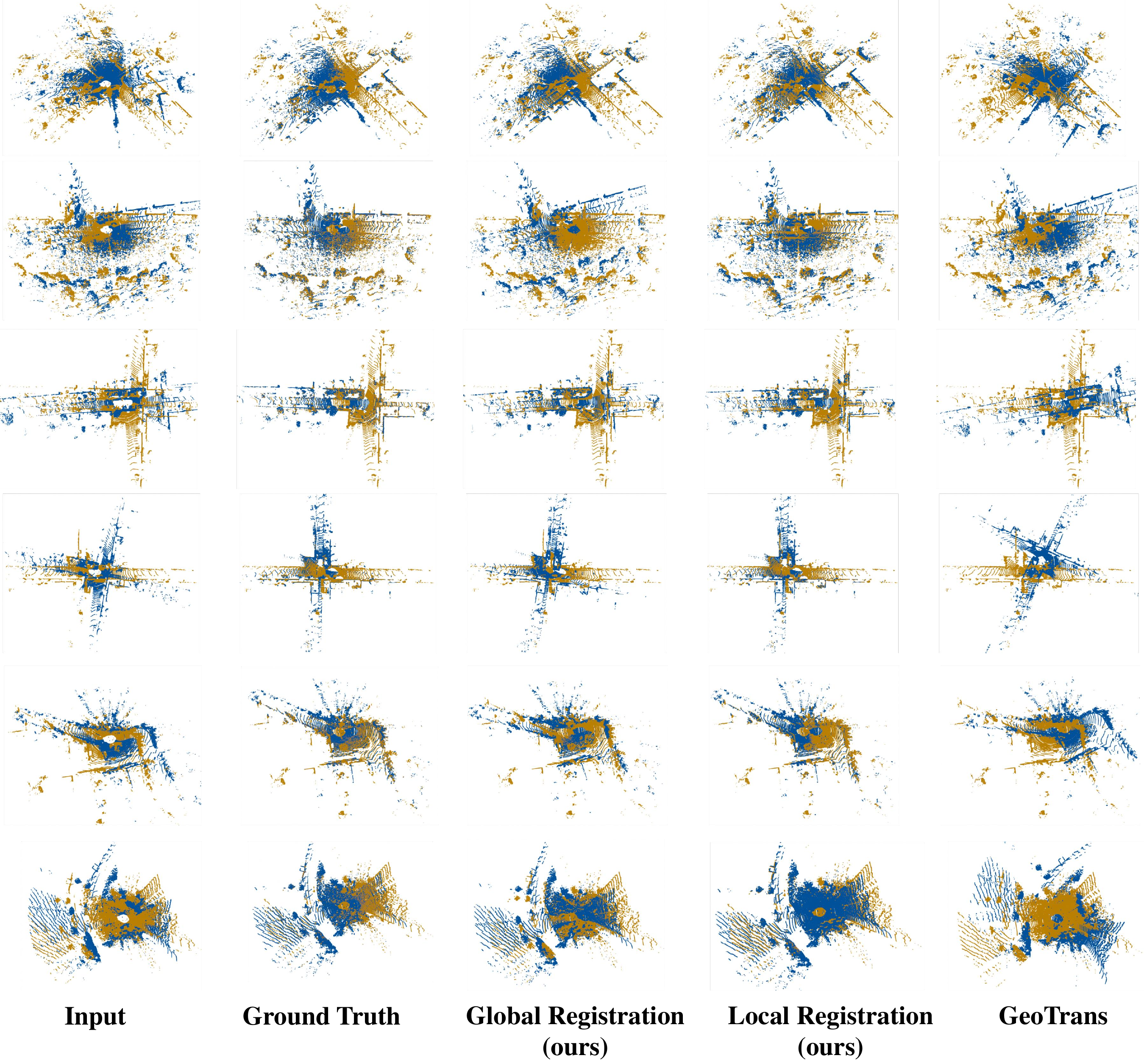}
    % \captionsetup{justification=justified, singlelinecheck=false}
\caption{Registration results on KITTI. Our method achieves higher accuracy after local registration and achieves good performance similar to ground truth.}
\label{fig:supp_result kitti}
\end{figure*}

For 3DMatch/3DLoMatch, RR is computed based on the Root Mean Square Error (RMSE), which is used to measure the distance error between two point clouds. For the set of ground truth correspondences $C$ after applying the estimated transformation $T_{est}$, the calculation formula of RMSE is as follows:
\begin{equation} 
RMSE = \sqrt{\frac{1}{|C|}{\sum_{(p_{i},q_{i}) \in C}||T_{est}(p_{i})-q_{i}||^{2}}} 
\end{equation} 
where $|C|$ is the number of correspondence set, $p_{i}$, $q_{i}$ are the i-th pair of paired points in the set of ground truth correspondences $C$. Lower RMSE values correspond to better alignment accuracy. RR is defined as:
\begin{equation}
RR = \frac{1}{|C|} \sum_{i=1}^{C}[RMSE_{i} < \tau_{RMSE}]  
\end{equation} 
where $\tau_{RMSE}$ is $0.2m$.
For KITTI, RR is defined as the ratio of point cloud pairs for which both the Relative Rotation Error (RRE) and Relative Translation Error (RTE) are below specific thresholds (i.e., RRE $< 5^\circ$ and RTE $< 2m$):
\begin{equation} 
RR = \frac{1}{|C|}\sum_{i}^{C}[RRE_{i} < 5^\circ \wedge RTE_{i} < 2m]
\end{equation} 

%%%%%%%%%%%%%%%%%%%%%%%%%%%%%%%%%%%%%%%%%%%%%%%%%

\section{Additional Experiment Results}
We also perform an ablation study on the 3DMatch dataset, as shown in the Table~\ref{tab:supp_ablation}. Similar to the experiment on the 3DMatch dataset in the main paper, we verify each block of our model separately and adopt the same experimental settings. The results prove that our default settings achieve the best performance.
Desides, we provide the registration results on KITTI in the Figure~\ref{fig:supp_result kitti} to show that our method achieves higher accuracy on the KITTI dataset.

\end{document}